\documentclass[twocolumn]{article}
\usepackage[T1]{fontenc}
\usepackage[margin=1in]{geometry}
\usepackage{authblk}
\usepackage{graphicx}
\usepackage{comment}
\usepackage{xcolor}
\usepackage{amsmath}
\usepackage{adjustbox}
\usepackage[utf8]{inputenc}

\usepackage{listings}
\newcommand{\promptline}[2]{%
  \noindent\textbf{#1:}\enspace\lstinline!#2!\\[-2pt]
}

\title{Evaluating ChatGPT’s Performance in Classifying Pneumonia from Chest X-Ray Images}

\author[1]{Pragna Prahallad}
\author[1]{Pranathi Prahallad}
\affil[1]{Emerald High School, Dublin, California\\
\texttt{pragnaprahallad@gmail.com}, \texttt{pranathiprahallad@gmail.com}}

\begin{document}
\maketitle

\begin{abstract}
In this study, we evaluate the ability of OpenAI’s \texttt{gpt-4o} model to classify chest X-ray images as either \texttt{NORMAL} or \texttt{PNEUMONIA} in a zero-shot setting, without any prior fine-tuning. A balanced test set of 400 images (200 from each class) was used to assess performance across four distinct prompt designs, ranging from minimal instructions to detailed, reasoning-based prompts. The results indicate that concise, feature-focused prompts achieved the highest classification accuracy of 74\%, whereas reasoning-oriented prompts resulted in lower performance. These findings highlight that while ChatGPT exhibits emerging potential for medical image interpretation, its diagnostic reliability remains limited. Continued advances in visual reasoning and domain-specific adaptation are required before such models can be safely applied in clinical practice.
\end{abstract}

\section{Pneumonia - A Health Challenge}
Pneumonia is an infection that causes inflammation in the air sacs of the lungs, called alveoli, and can occur in one or both lungs. The air sacs in the lungs can fill with fluid or pus, resulting in coughs accompanied by phlegm or pus. The filling of the alveoli (air sacs) can make it harder for oxygen to reach the bloodstream\cite{mayoclinic_symptoms}.

Pneumonia can be life-threatening to infants,  people over the age of 65, and those with weakened immune systems. Viruses, bacteria, fungi, and bacteria-like organisms are all leading causes of pneumonia. Additionally, there are three other ways pneumonia can be acquired: Hospital-acquired pneumonia, Healthcare-acquired pneumonia, and aspiration pneumonia. Key symptoms of pneumonia include chest pain when breathing or coughing, fatigue, sweating, fever, body chills (signs of infection), nausea, vomiting, diarrhea, and shortness of breath, confusion or changes in mental awareness, and lower than normal body temperature in those who are over the age of 65. 

Pneumonia remains one of the biggest health challenges in the world. In the United States alone, 1.4 million emergency room visits were a result of pneumonia. Additionally, over 40,000 deaths were due to pneumonia \cite{lungassociation_pneumonia}. According to the World Health Organization (WHO), in 2017, pneumonia killed over 808,000 children under the age of 5, accounting for 15 percent of deaths in children under 5 ~\cite{who_pneumonia} . The high mortality rates of pneumonia tie back to delayed diagnoses, lack of equipment in low-resource areas, and the difficulty of differentiating pneumonia from other respiratory conditions. These statistics emphasize the need for a faster, more efficient, and cost-effective method to diagnose pneumonia, an area that requires more research and testing, including in AI-assisted imaging. 

\subsection{Detection of Pneumonia in Clinics}
Pneumonia is most commonly detected by a doctor asking for a medical history and using a stethoscope to listen for bubbling or crackling sounds when placed on your lungs. If a doctor suspects pneumonia, there are four main tests to pinpoint the cause of the infection and prevent its spread. First, a blood test can be used to confirm the presence of an infection and attempt to identify the type of organism (virus, fungus, bacteria, or bacteria-like organisms) causing the infection.\cite{mayoclinic_diagnosis} A drawback of blood tests is that they are not specialized in precise identification. Second, a doctor might conduct a chest X-ray, which can help determine the extent and location of the pneumonia. Third, a pulse oximetry can be conducted to measure the level of oxygen in the blood. Lastly, a sputum test can aid in zeroing in on the cause of infection.

\section{The Aid of AI in Detecting Pneumonia}
Fast, accurate, and efficient diagnoses of pneumonia are essential for patient health, especially for patients over the age of 65. A misdiagnosis of pneumonia can have devastating effects on their health, leading to long-term complications. A study titled “Inappropriate Diagnosis of Pneumonia Among Hospitalized Adults”, twelve percent of all pneumonia diagnoses are misdiagnosed~\cite{gupta2024pneumonia}. AI has the potential to aid in faster, more consistent interpretation of chest scans and X-rays.

\subsection{Zero-Shot Learning for Image Classification}
Zero-shot learning (ZSL) is a paradigm in machine learning where a model is capable of recognizing and classifying new categories that it has never encountered during training. Unlike conventional supervised models that rely on large labeled datasets, ZSL leverages the model’s ability to associate visual inputs with semantic or textual representations learned during pretraining \cite{liu2023chatgptzero}. In the context of medical imaging, zero-shot learning allows an AI model—such as a large vision-language model—to analyze radiographs or scans and infer diagnostic labels based on semantic understanding, rather than direct exposure to labeled examples of each condition.

\subsection{The Role of Prompts in Zero-Shot Inference}
In zero-shot image classification using large multimodal models, prompts act as the interface between human intent and model behavior. They define the task, provide linguistic context, and specify the expected output format. Effective prompt design aligns the model’s general visual–linguistic knowledge with the target task, emphasizing relevant features while reducing ambiguity.

Prompts vary in their level of specificity. Some may request a concise classification, while others guide the model to explain its reasoning or use domain-specific terminology. By carefully structuring these instructions, prompts enable zero-shot models to transfer learned knowledge to new domains without explicit supervision.

Furthermore, prompts can influence the model’s reasoning process. Including instructions to focus on key diagnostic features or to provide step-by-step explanations can improve interpretability and reduce errors arising from incomplete or inconsistent reasoning. Overall, prompt design plays a critical role in directing

\subsection{Image Classification in Health Care}
Recent research has increasingly explored the integration of zero-shot learning into medical diagnosis and analysis. A 2025 study in the \textit{Journal of Medical Internet Research} demonstrated how prompt engineering, chain-of-thought reasoning, and retrieval-augmented generation can enhance diagnostic accuracy for pulmonary diseases~\cite{zhang2025pulmonary}. Although this work focused on textual radiology reports rather than images~\cite{zhang2023healthprompt}, it emphasized how prompt design and reasoning strategies directly influence model performance and interpretability.

Building on this, \textit{HealthPrompt: A Zero-Shot Learning Paradigm for Clinical Natural Language Processing} introduced a framework for classifying medical texts through prompt-based learning, further underscoring the importance of linguistic structure and task formulation in achieving accurate predictions.  

In the visual domain, studies have extended zero-shot and few-shot prompting methods to medical imaging tasks. For instance, \textit{Improving Pneumonia Localization via Cross-Attention on Medical Images and Reports} leveraged both image and textual features to enhance pneumonia detection~\cite{zhao2021pneumonia}. Similarly, the \textit{PM2: Prompting Multi-Modal Model for Few-Shot Medical Image Classification} framework combined textual and visual cues for improved classification accuracy~\cite{li2024pm2}. More recently, \textit{A ChatGPT-Aided Explainable Framework for Zero-Shot Medical Image Classification} utilized large language models such as ChatGPT to directly interpret and classify medical images, highlighting the growing potential of zero-shot approaches in healthcare~\cite{chen2024chatgpt}.


\subsection{Previous Work in Pneumonia Image Classification} 
Machine learning has played a pivotal role in automating medical image diagnosis, including the detection of pneumonia from chest X-rays. Early studies, such as that by Mujahid et al.~\cite{mujahid2022pneumonia_inception}, employed Inception-based architectures to distinguish between normal and pneumonia-infected lungs. Their work demonstrated that transfer learning could achieve high accuracy even with limited training data. Similarly, El Asnaoui and Chawki~\cite{elAsnaoui2021_design_ensemble} developed an ensemble deep learning framework to improve reliability and minimize classification errors. Kundu et al.~\cite{kundu2021_ensemble_transfer} further showed that combining multiple deep learning models enhanced both recall and robustness compared to individual networks.

More recently, Singh et al.~\cite{singh2024_viT_pneumonia} introduced Vision Transformers (ViTs) for pneumonia detection, leveraging attention mechanisms to focus on critical image regions. Mabrouk et al.~\cite{mabrouk2022_ensemble_pneumonia} extended this approach by integrating ViTs with other deep learning models, achieving further improvements in classification accuracy.

Despite these advances, most existing approaches rely on fully supervised learning with large labeled datasets. In contrast, emerging large multimodal models, such as ChatGPT with vision capabilities, can interpret and classify medical images through natural-language prompts without task-specific retraining. This shift introduces new possibilities for zero-shot image classification in healthcare. 

In this study, we investigate the use of OpenAI models for classifying chest X-rays as normal or pneumonia-affected using zero-shot learning guided by prompt engineering, and we evaluate their diagnostic accuracy.

\section{Experimental Methodology} 
\subsection{Database Used}
This study used the publicly available \textit{Chest radiographic Images (Pneumonia)} data set from Kaggle~\cite{mahmoud2018chestxray}, which is widely used to benchmark pneumonia detection models. The data set consists of a total of 5,863 anterior chest X-ray images, categorized into two classes: \textit{Normal} and \textit{Pneumonia}. The images were taken from pediatric patients aged one to five years at the Guangzhou Women and Children’s Medical Center, China.

The data set is organized into three subsets:
\begin{itemize}
    \item \textbf{Training set:} 5,216 images
    \item \textbf{Test set:} 624 images
    \item \textbf{Validation set:} 16 images
\end{itemize}

Among the 624 test images, 234 show normal chest radiographs, while 390 show radiographic evidence of pneumonia. Since the goal of this work was to evaluate the performance of zero-shot image classification, only the \textbf{test set} was utilized in our experiments. All images were used in their original form without fine-tuning or supervised training, aligning with the zero-shot learning framework.

\subsection{Sample Chest X-Ray Images}
\begin{figure}[h!]
    \centering
    \includegraphics[height=4cm]{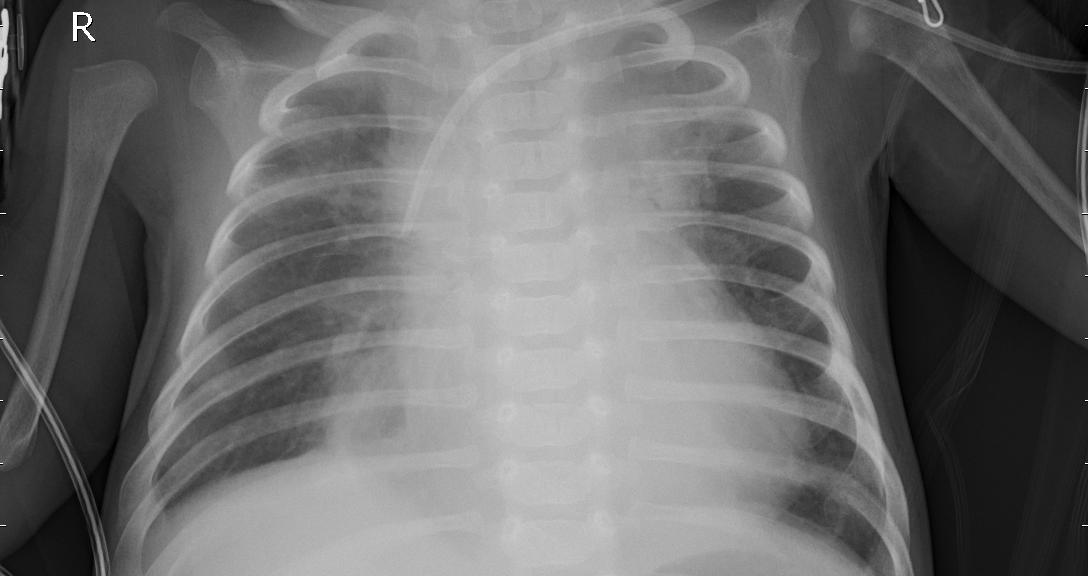}
    \hspace{0.5cm}
    \includegraphics[height=4cm]{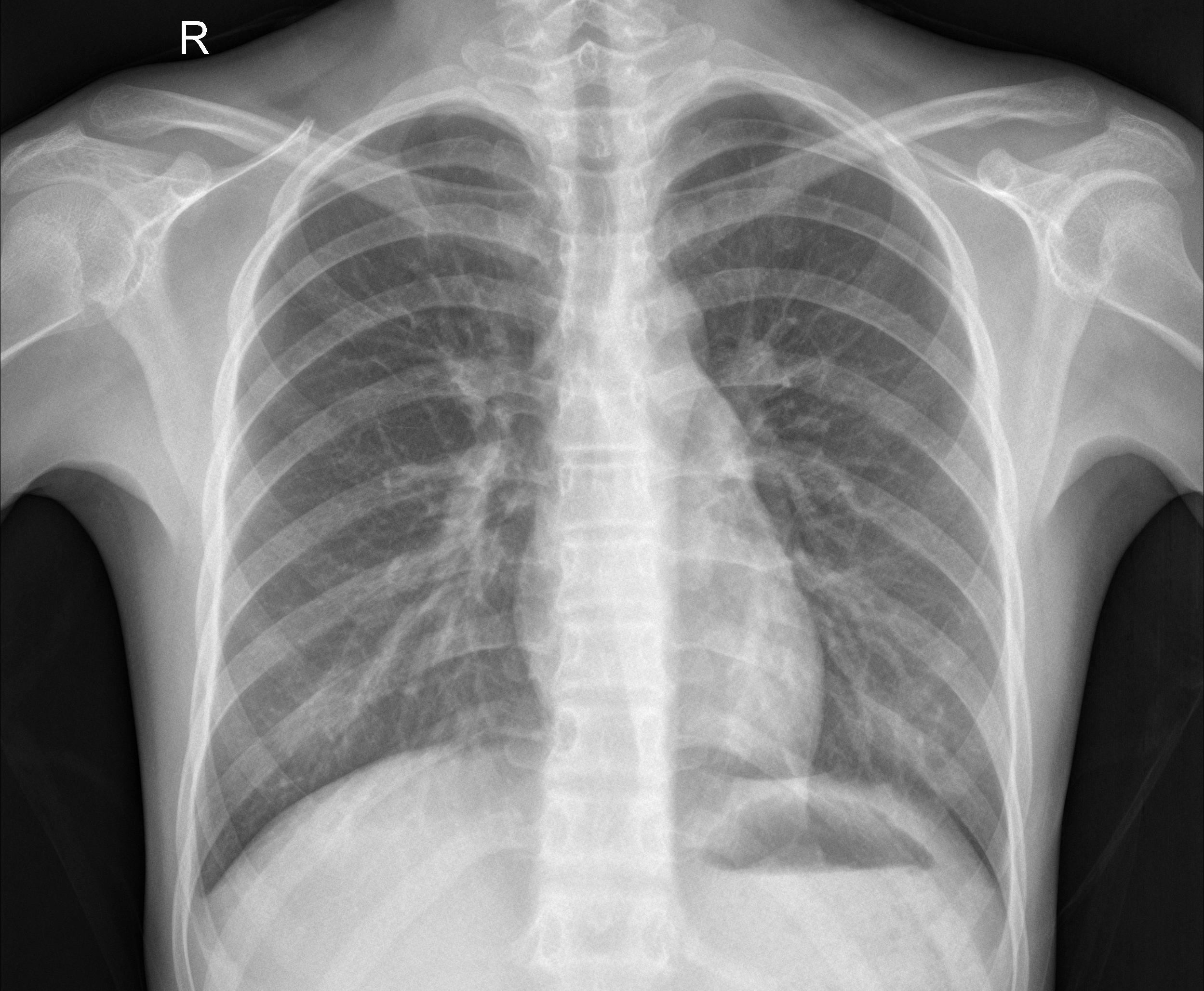}
    \caption{Example chest X-ray images from the test set in the \textit{Chest X-Ray Images (Pneumonia)} dataset~\cite{mahmoud2018chestxray}. 
    (a) Pneumonia X-ray \hspace{1cm} (b) Normal X-ray.}
    \label{fig:xray_samples}
\end{figure}

To illustrate the dataset, Fig.~\ref{fig:xray_samples} shows two representative chest X-ray images from the test set. The first image corresponds to a pneumonia-affected lung, while the second image represents a normal chest X-ray. These examples highlight the visual contrast between infected and healthy lungs that the model aims to classify.

\section{Model and Prompt Design}

\subsection{Model and Image Preprocessing}

We used the \texttt{gpt-4o} model (also called ``GPT-4 Omni''), a version of ChatGPT that can understand both pictures and text together. This means that it can look at a chest X-ray while reading written instructions (called prompts) to decide if the image shows a healthy chest (\texttt{NORMAL}) or one with pneumonia (\texttt{PNEUMONIA}).

Before each X-ray was sent to the model, the images were prepared so that they were clear but not too large. Each image was opened in color mode, resized so that its longest side was no more than 2048 pixels, and saved in high quality JPEG format. This ensured that all images had the same quality and size, helping the model analyze them fairly and accurately.

\subsection{Prompt Design}

To communicate effectively with the model, each interaction used two types of messages: a \textbf{system prompt} and a \textbf{user prompt}.  

The \textbf{system prompt} tells the model how to behave or what role it should take—for example, to act like a medical image labeler and to give answers only in JSON format. It sets the overall tone and rules for the task.  

The \textbf{user prompt}, on the other hand, gives the actual instruction for that specific image—for example, what kind of answer to return, what format to use, and the image itself. Together, the system and user prompts help the model stay consistent, structured, and focused while analyzing the X-rays.

We tested four prompt types to see how the way we ask the model affects its answers. Each prompt gave the model both the X-ray image and a short instruction. The model always returned results in strict JSON format. The four prompt types are shown below.

\subsubsection{Prompt 1: Minimal Output (No Features)}

\promptline{System}{You are labeling chest X-rays as either NORMAL or PNEUMONIA. Output strict JSON only.}
\promptline{User}{Return ONLY JSON like this: {"label":"NORMAL|PNEUMONIA","confidence":0..1}. (Image attached)}

This was the simplest version. The model only had to pick a label and confidence value, without explaining its choice.

\subsubsection{Prompt 2: Including Features in Output}

\promptline{System}{You are labeling chest X-rays as either NORMAL or PNEUMONIA. Output strict JSON only.}
\promptline{User}{Return ONLY JSON like this: {"features":"...","label":"NORMAL|PNEUMONIA","confidence":0..1}. (Image attached)}

This version asked the model to also list short visual features it noticed in the image, such as “cloudy lung areas” or “clear air spaces.”

\subsubsection{Prompt 3: Features + Concise Reasoning}

\promptline{System}{You are labeling chest X-rays for dataset use. Be objective and return strict JSON only. Do not include step-by-step reasoning; just give a short justification.}
\promptline{User}{Return ONLY JSON with keys: {"features":"<=160 chars","reason":"<=240 chars (short justification)","label":"NORMAL|PNEUMONIA","confidence":0..1}. (Image attached)}

This version added a short reason (one or two sentences) to explain why the model made its choice, while staying brief and direct.

\subsubsection{Prompt 4: Features + Step-by-Step Reasoning}

\promptline{System}{You are labeling chest X-rays for dataset use. Be objective and return strict JSON only. Include step-by-step reasoning.}
\promptline{User}{Return ONLY JSON with keys: {"features":"<=160 chars","reason":"<=512 chars (step-by-step)","label":"NORMAL|PNEUMONIA","confidence":0..1}. (Image attached)}

This final version asked the model to describe its reasoning process in more detail before deciding.

\subsection{Summary of Results}

Each prompt was tested on 400 X-rays (200 labeled as \texttt{NORMAL} and 200 as \texttt{PNEUMONIA}). The results are summarized in Table~\ref{tab:prompt_results}.

\begin{table}[h]
\centering
\caption{Accuracy Comparison of Prompt Variants}
\label{tab:prompt_results}
\begin{tabular}{lccc}
\hline
\textbf{Description}&\textbf{Acc.(\%)} \\
\hline
Minimal Output (No features) & 64.50  \\
With feature output & 74.00  \\
Features + concise reasoning & 71.50  \\
Features + step-by-step reasoning & 70.75  \\
\hline
\end{tabular}
\end{table}

\noindent
Adding short visual features (Prompt 2) helped the model better identify \texttt{NORMAL} images. However, asking for longer reasoning (Prompt 4) did not improve results further, suggesting that brief, focused reasoning works best for image-based classification tasks.

\section{Conclusion}

This study demonstrates that while ChatGPT (\texttt{gpt-4o}) is capable of classifying chest X-rays into \texttt{NORMAL} and \texttt{PNEUMONIA} using carefully designed prompts, its diagnostic accuracy remains limited. Even with the best-performing prompt, the overall accuracy reached only 74\%, indicating that the model still struggles to consistently recognize subtle medical image features. 

The results also show that prompts requiring longer or step-by-step reasoning did not improve performance—in fact, they slightly reduced accuracy. This suggests that reasoning-oriented large language models, while powerful for text-based tasks, are not yet optimized for visual diagnostic reasoning~\cite{jiang2024gpt4vgenerateradiologyreports}. In image-based classification, especially in medical settings, concise and feature-focused prompts appear more effective than reasoning-heavy instructions~\cite{liu2025mindstepbystep}.

Overall, this experiment highlights the potential of multimodal models like ChatGPT in medical imaging, but also emphasizes the need for future improvements in visual understanding, specialized medical fine-tuning, and integration with domain-specific datasets to achieve clinical reliability.



\bibliographystyle{ieeetr}
\bibliography{references}
\end{document}